\documentclass[letterpaper, 10 pt, conference]{ieeeconf}  

\IEEEoverridecommandlockouts                              

\usepackage{siunitx}
\usepackage{times}
\usepackage{xcolor}
\usepackage{graphicx}
\usepackage{amsmath}
\usepackage{amsfonts}
\usepackage{placeins}
\usepackage{multicol}
\usepackage{multirow}
\usepackage{booktabs}
\usepackage{xspace}
\usepackage[bookmarks=true]{hyperref}
\usepackage{xcolor}
\usepackage{colortbl}

\makeatletter
\long\def\@makecaption#1#2{%
\ifx\@captype\@IEEEtablestring
  \parbox[t]{\hsize}{\normalsize \emph{#1}.~~#2}%
  \@IEEEtablecaptionsepspace
\else
  \@IEEEfigurecaptionsepspace
  \setbox\@tempboxa\hbox{\normalsize #1.~~ #2}%
  \ifdim \wd\@tempboxa >\hsize
    \setbox\@tempboxa\hbox{\normalsize #1.~~ }%
    \parbox[t]{\hsize}{\normalsize\noindent\unhbox\@tempboxa#2}%
  \else
    \ifcenterfigcaptions \hbox to\hsize{\normalsize\hfil\box\@tempboxa\hfil}%
    \else \hbox to\hsize{\normalsize\box\@tempboxa\hfil}%
    \fi
  \fi
\fi}
\makeatother

\definecolor{ourscolor}{HTML}{2471A3}   
\colorlet{oursrow}{ourscolor!16!white}

\newcommand{\mpc}{\textbf{(\ref{item:mpc})}\xspace}
\newcommand{\rlonlyrealobs}{\textbf{(\ref{item:rlonlyrealobs})}\xspace}
\newcommand{\rlgtobsnotcc}{\textbf{(\ref{item:rlgtobsnotcc})}\xspace}
\newcommand{\rlgtobsteacher}{\textbf{(\ref{item:rlgtobsteacher})}\xspace}
\newcommand{\daggerrealobsstudent}{\textbf{(\ref{item:daggerrealobsstudent})}\xspace}

\definecolor{projecturlcolor}{HTML}{39ACF8}

\definecolor{projecturlcolor}{HTML}{1A4DB5} 
\newcommand{\projecturlbare}{https://arth-shukla.github.io/catch-it-all}
\DeclareRobustCommand{\projecturl}{%
  \href{\projecturlbare}{\textcolor{projecturlcolor}{\urlstyle{tt}\textbf{\nolinkurl{\projecturlbare}}}}\xspace}

\title{
\emph{Catch It All!}: Generalizable Dynamic Catching with a Dexterous Hand
}

\vspace{-0.5in}

\author{Arth Shukla$^{*,1}$, Kehlani Fay$^{*,1}$, Huaze Liu$^2$, Darin Djapri$^1$, Alexiy Buynitsky$^1$, Nicklas Hansen$^1$ \\
$^1${UC San Diego}\qquad$^2${Harvey Mudd College}\qquad$^*${Equal contribution} \\
\projecturl}

\begin{document}

\IEEEaftertitletext{%
\begin{center}
    \vspace{-0.3in}
    \includegraphics[width=\linewidth]{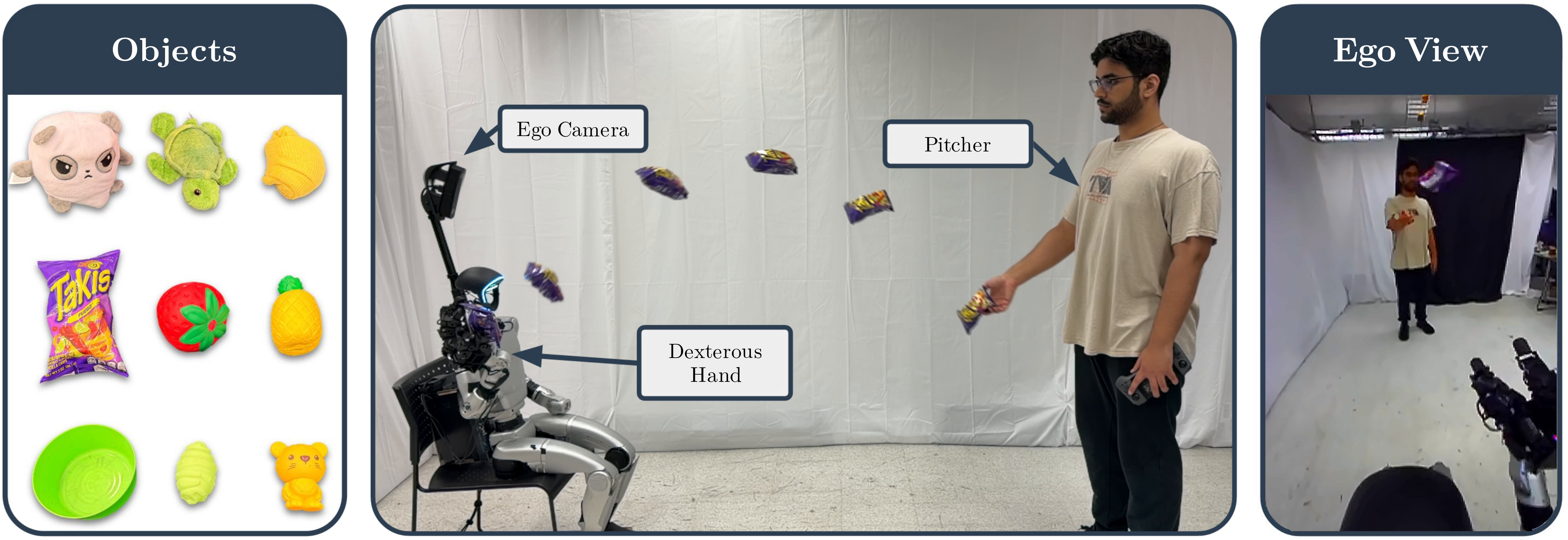}
    \vspace{-0.1in}
    \refstepcounter{figure}\label{fig:coverish}
    \parbox{\textwidth}{\normalsize
    \text{Fig.~\thefigure.}{ }{ }
    \textbf{Overview.} Our method successfully catches diverse, unseen objects overhand and underhand using only an egocentric camera. Left: The real-world object test set includes 9 unseen objects with variable size, elasticity, deformability, and shape. Middle: Catching setup with the robot positioned overhand, and the human tossing an object selected at random from the test set. Right: Egocentric view; this is the only camera in our setup.
    }
\end{center}
}

\maketitle
\thispagestyle{empty}
\pagestyle{empty}

\begin{abstract}

Catching arbitrary objects requires generalization to novel projectiles and real-time adaptation to unpredictable contact dynamics under noisy visual observations. This paper presents \emph{Catch It All!}, a sim-to-real catching framework for humanoid robots that generalizes and adapts to unseen objects with variable physical properties for underhand and overhand catches, using only one egocentric camera. We learn policies which generalize to unseen objects with varying geometry, deformability, elasticity, and physical properties. Effective grasps which adapt real-time to object poses and shapes are learned through careful grasp reward design. We provide safety-aware curricula which enable safe, high-frequency action execution near hardware limits, including wrist torque limits as low as \SI[inter-unit-product=\ensuremath{{\cdot}}]{5}{\newton\meter}. We validate our system by deploying zero-shot on a Unitree G1. With only one egocentric camera, \emph{Catch It All!} successfully performs underhand and overhand catches on 9 unseen objects not seen during training, including low-rebound plushes, semi-deformable toys and chip bags, and rigid-elastic objects with high rebound on contact.

\end{abstract}

\section{INTRODUCTION}

Dexterous catching is a fundamental dynamic manipulation problem, requiring precise synergy between vision and control under tight timing constraints ~\cite{gold2022catching,kober2012playing}. For a successful catch, robots must achieve low-latency, adaptable, and generalizable visuomotor reflexes, as thrown objects vary widely in their contact and grasp dynamics. On humanoid platforms, this problem is compounded. Unlike industrial manipulators, humanoid arms typically operate under significantly lower torque limits and control bandwidth, making it easy to exceed hardware limits and damage a robot. Humanoids rely on lighter actuators to ensure safety and work within mobile power constraints. Catching requires learning high-frequency, reactive behaviors without exceeding these hardware limits. Prior systems demonstrate high-speed interception on quadrupeds and industrial manipulators~\cite{gold2022catching,forrai2023event,schakkal2024dynamic,huang2023dynamic}; unlike these platforms, with wrist torque limits as low as \SI[inter-unit-product=\ensuremath{{\cdot}}]{5}{\newton\meter}, humanoids are more susceptible to hardware limits during high-frequency action execution. 

Generalizable catching across diverse objects with a wide range of physical properties is a challenge. Prior sim-to-real works rely on simplified projectiles with low elasticity such as sandbags~\cite{huang2023dynamic, zhang2024catch} which are more forgiving to imprecise timing or weak grasps. In simulation, prior work has achieved catching diverse projectiles~\cite{lan2025dexcatch}; however, zero-shot transfer to general unseen objects in the real world remains an open problem. To overcome these limitations, our framework leverages careful reward and curriculum design, achieving zero-shot real world transfer with generalization to unseen objects while executing safely within hardware limits.

We present \textbf{\emph{Catch It All!}}, a generalizable dexterous catching framework for underhand and overhand catching that operates on a single egocentric camera without motion capture or human demonstrations. Our contributions are: \textbf{(1)} generalizing to novel objects in both underhand and overhand catches through grasp reward design and a toss curriculum, \textbf{(2)} safe action execution near hardware limits via curricula on torque clipping and reward penalties, and \textbf{(3)} zero-shot sim-to-real transfer, capable of catching objects unseen in simulation both underhand and overhand.



\begin{figure*}[t!]
    \centering
    \includegraphics[width=\textwidth]{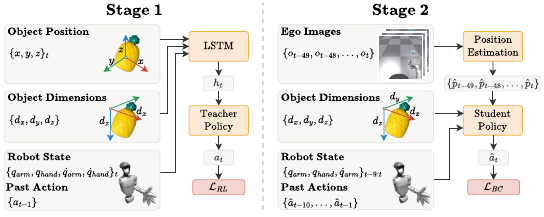}
    \vspace{-0.3in}
    \caption{\textbf{Teacher-student training.} \emph{Stage 1:} An RL teacher policy is trained on ground-truth projectile position. \emph{Stage 2:} A student policy imitates the frozen teacher via DAgger using noisy camera observations and a $50$-step (1 second) history.}
    \vspace{-0.1in}
    \label{fig:method}
\end{figure*}

\section{RELATED WORK}

\subsection{Challenges in Robotic Catching}
Catching requires a robot to rapidly adapt control outputs to noisy, evolving visual estimates of an object's state. The robot must estimate the object's state, predict its trajectory, and plan and execute body control within milliseconds~\cite{kim2014catching, baeuml2010kinematically, lampariello2011trajectory}. With an egocentric camera, the object may also leave the field of view during flight, making continuous state estimation more difficult. These estimates cannot be treated as fixed because new, noisy observations continually update and affect both the predicted trajectory and the required control correction. For catching, the controller must adapt and correct its actions rapidly online.

This challenge is compounded by the fact that the object's motion depends on dynamics that are often unknown in advance. Properties such as spin~\cite{zhang2014spin} can substantially affect the predicted trajectory, so a model tuned to one object offers little guarantee of generalizing to another. The same uncertainty extends to selecting a stable catch configuration, which depends on the object's mass, center of mass, and inertia. These properties must be estimated or learned online, under uncertainty, prior to contact~\cite{mavrakis2020estimation}. At impact, sensorimotor imprecision may cause the fingers to close too abruptly, letting the object escape the hand before the grasp is complete~\cite{salehian2016dynamical}. The robot's grasp must also quickly adapt to the object's geometry and compliance to remain stable. Due to these complexities, catching remains an open problem.

\subsection{Generalization Limitations}
Prior work has made catching tractable by reducing task difficulty at the cost of task generalization. This has included simplifying catching by using a predefined set of known objects with dynamics fitted individually per object~\cite{kim2014catching, salehian2016dynamical}. However, this limits generalizability to unseen objects under varying contact dynamics and physical properties. Similarly, prior work has leveraged precise external motion capture systems rather than egocentric sensing~\cite{kim2014catching, salehian2016dynamical, dong2020catch}. Egocentric vision must handle high noise characteristics of a camera mounted on a shifting robot, reducing state estimation accuracy of the object. 

Other approaches simplify the object itself for simpler contact dynamics. For instance, restricting catching to sandbags, which exhibit low rebound and are more forgiving of tracking error~\cite{zhang2024catch}, or to standard balls with no variable shape~\cite{dong2020catch, bauml2011catching, frese2001offtheshelf}. In the context of multi-agent catching,~\cite{huang2023dynamic} reduces the physical toss range to a short handover between two close robot arms trained in a multi-agent setting. While this requires learning coordination between arms, it also limits the diversity of tosses the catching arm is capable of capturing. Others leverage simpler manipulators to reduce the search space and degrees of freedom, such as nets mounted on quadrupeds~\cite{forrai2023event} or a standing quadruped's front legs~\cite{schakkal2024dynamic}. 

Moreover, extending these capabilities to humanoid platforms introduces additional challenges: compared to industrial manipulators, humanoid actuators typically exhibit lower torque limits and control bandwidth~\cite{hwangbo2019learning,rudin2022learning}. Enabling dynamic dexterous manipulation on such platforms requires safe action execution at hardware limits. These simplifications in object diversity, vision, toss range, and manipulator dexterity collectively limit prior systems to narrow, controlled catching scenarios, motivating the need for an approach that catches diverse, previously unseen objects using only onboard, egocentric perception.

\section{METHOD}

\begin{figure*}[ht!]
    \centering
    \includegraphics[width=\textwidth]{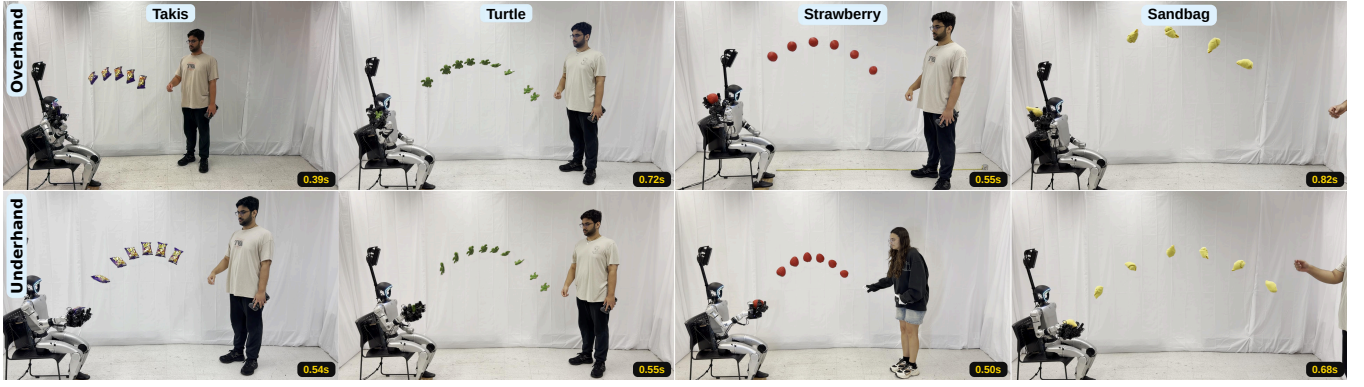}
    \vspace{-0.2in}
    \caption{\textbf{Qualitative Results.} Examples of successful catches across 4 of the 9 real-world objects: Takis, turtle, strawberry, and sandbag. Top, overhand: policy catches short tosses by moving the hand forward, while it catches long tosses by moving the hand backwards mid-capture. Bottom, underhand: the hand tracks the object during the toss to intercept and grasp it.}
    \label{fig:qualitative}
\end{figure*}

\subsection{System Setup}
\label{sec:sys_setup}

Compared to industrial systems, humanoids often have lower actuator bandwidth and coarser control resolution. Learning high-frequency, reactive control while operating near the arm's hardware limits remains a roadblock for performing dynamic dexterous manipulation tasks, such as catching with humanoids. Thus, we choose our setup to tackle sim-to-real transfer for dexterous catching with a humanoid arm.

For the robot catching setup, we deploy on a seated Unitree G1 EDU with a ZED2i mounted overhead for full view of egocentric observations. We use the 20-DoF House of Dextra hand~\cite{fay2025crossembodied}. Compared to the 16-DoF LEAP Hand (\SI{595}{g})~\cite{leaphand} and 6-DoF Inspire RH56BFX (\SI{540}{g}), the House of Dextra hand provides 4 additional DoF through an extra finger, with comparable mass (\SI{600}{g}), and a smaller profile than LEAP. The high DoF allows for rapid adaptation with diverse grasps across objects with varying size and shape (see Fig. \ref{fig:strong_grasps} for examples of grasps from real-world tosses). We additionally print a hollow, carbon-fiber palm to ensure the hand remains within the G1 arm's nominal \SI{3}{kg} payload.

\begin{figure}[t]
    \centering
    \includegraphics[width=\linewidth]{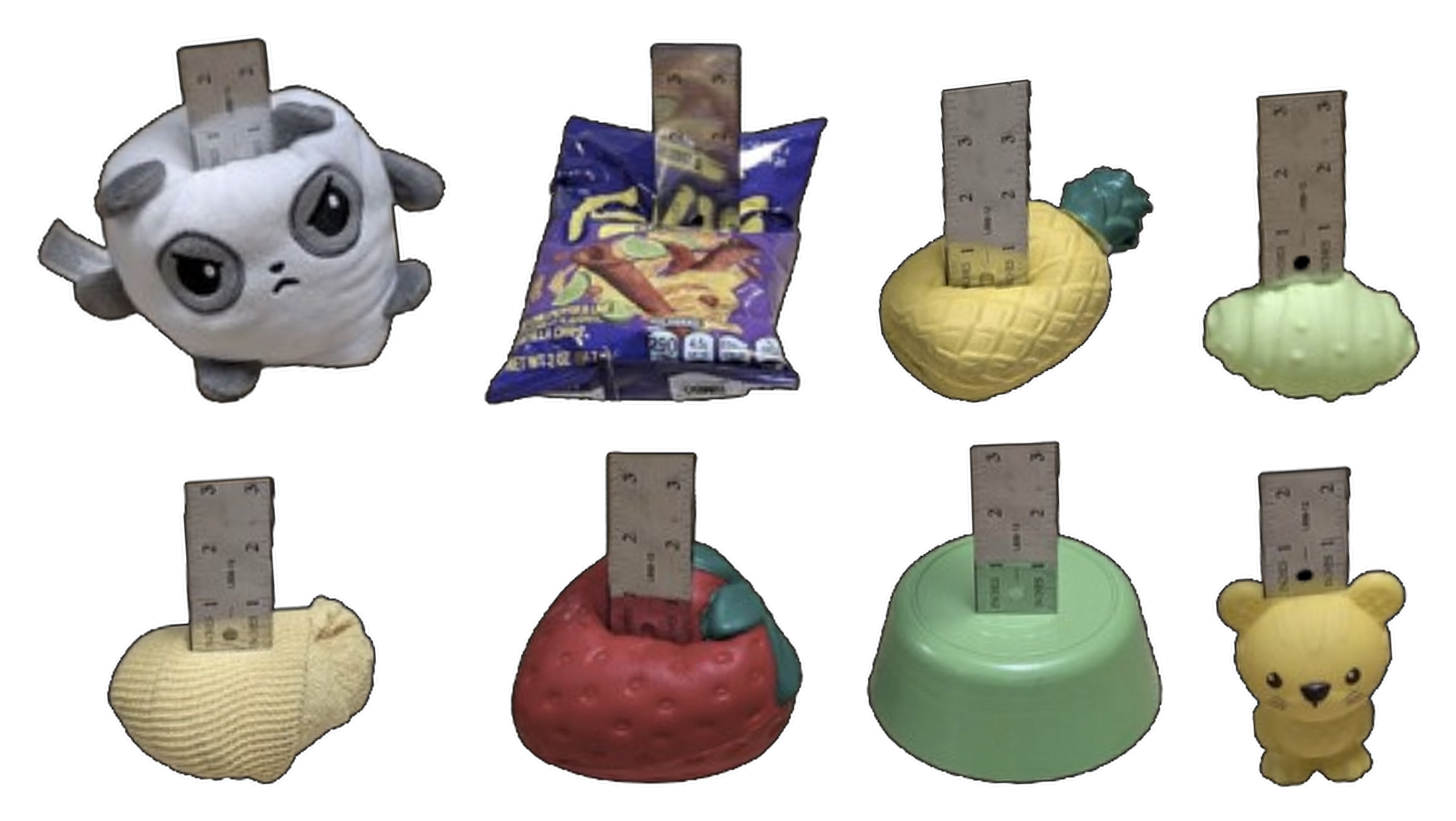}
    \vspace{-0.2in}
    \caption{\textbf{Real-world tossing objects.} Real-world projectiles unseen in training; we exert 1 pound of force to show variations in deformability, geometry, and physical properties.}
    \label{fig:realobjdeform}
\end{figure}

Moreover, while we train our policies in a rigid-body simulator, our real-world experiments also include semi-deformable objects. The deformability of real objects under 1 pound of force is pictured in Fig. \ref{fig:realobjdeform}.

\subsection{Policy Training}
\label{sec:policy_training}

We formulate our catching task as a Partially-Observable Markov Decision Process (POMDP) which can be described as tuple $\mathcal{M} = (\mathcal{S}, \mathcal{O}, \mathcal{A}, \mathcal{R}, \mathcal{T}, \rho, \gamma)$ with continuous state space $\mathcal{S}$ partially observed by observation space $\mathcal{O}$, action space $\mathcal{A}$, scalar reward function $\mathcal{R}: \mathcal{S} \times \mathcal{A} \rightarrow \mathbb{R}$, environment dynamics function $\mathcal{T}: \mathcal{S} \times \mathcal{A} \rightarrow \mathcal{S}$, initial state distribution $\rho$, and discount factor $\gamma \in [0, 1]$. The policy receives proprioception, 3D object position extracted with SAM2-Tiny~\cite{ravi2024sam2}, and 3D bounding box dimensions for the object. The arm is controlled with full delta 6D pose via Inverse Kinematics, and the hand with absolute joint control.

We train separate underhand and overhand policies. As seen in Sec. \ref{sec:robust_and_gen}, due to noisy stereo depth readings and dropped position prediction when objects leave the frame, exploration is challenging for purely online RL. So, as depicted in Fig. \ref{fig:method}, we adopt a teacher-student framework: teacher policies are trained with PPO~\cite{ppo} with ground-truth object position observations, then distilled to student policies with noisy observations using DAgger~\cite{dagger}.

To increase simulation throughput, we train on primitive shapes with simple meshes. We additionally evaluate on irregular objects and long-axis objects, depicted in Fig. \ref{fig:simobjs}. We randomize object friction, mass, scale, and restitution. Tosses are initialized at \SIrange{0.5}{2.75}{m} forward, \SI{\pm1.0}{m} lateral, and \SIrange{0.25}{2.0}{m} height, sampled at 40\% short, 40\% medium, and 20\% long range. We also randomize observation delay, control delay, toss delay, initial 6D hand pose, gravity, PD gains, link masses and friction, and random force injections.

We use IsaacGym~\cite{makoviychuk2021isaac} for policy training and Mujoco~\cite{mujoco} for sim-to-sim evaluations. Our training codebase extends HumanoidVerse~\cite{HumanoidVerse}. Policies are trained, evaluated, and deployed on a single workstation with a Ryzen 5 5600X CPU and RTX 4090 GPU.

\subsection{Rewards}

\label{sec:rewards}

\begin{figure}[t]
    \centering
    \includegraphics[width=\linewidth]{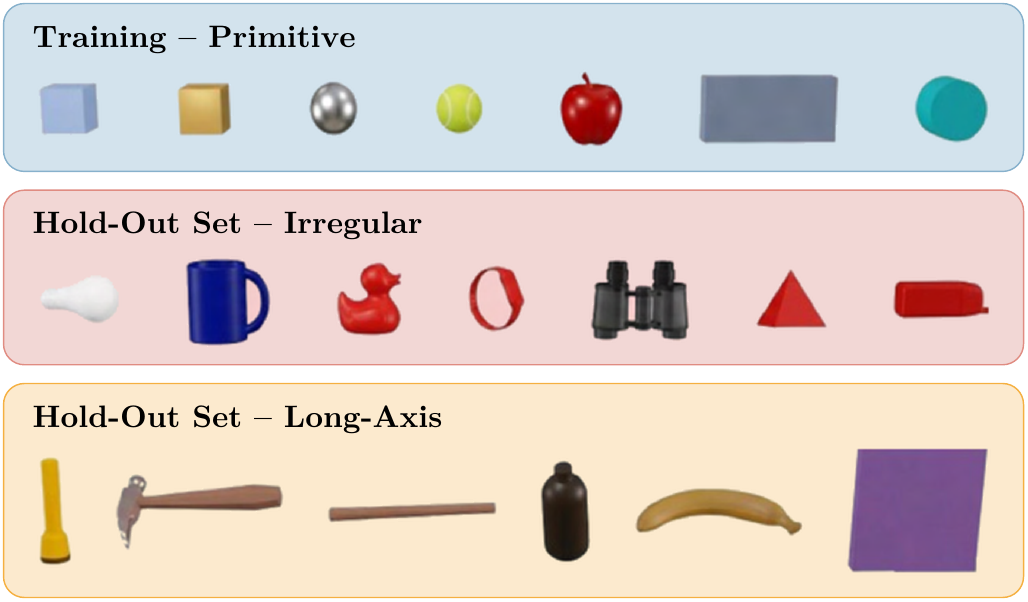}
    \vspace{-0.2in}
    \caption{\textbf{Simulation objects.} Top: training uses only primitive objects with simple meshes to increase throughput. Middle: hold-out set of irregular objects with novel geometries, unseen during training. Bottom: hold-out set of long-axis objects with high angular velocity, unseen during training.}
    \label{fig:simobjs}
\end{figure}

In order to generalize well to diverse, unseen objects, we learn policies with robust grasps to successfully capture objects and prevent them from bouncing or slipping out of the hand. In Sec. \ref{sec:reward_abl}, we show a key reward term is our fingertip-object surface distance reward, which we design to provide denser signal as fingers approach the object, or when all fingers cannot touch the object simultaneously (e.g. small/irregular objects). First, we approximate finger-surface distance with an estimated object radius $r_{\mathrm{obj}}$. Importantly, this increases simulation throughput compared to true mesh-link distance computation, and is a reasonable approximation given the simpler geometries of our training set of primitive objects. Then, reward increases as the maximum distance from any finger to the object surface decreases: $\exp(-10 ( \max\limits_{\mathrm{ftip}} \|p_{*}-p_{\mathrm{obj}}\|-r_{\mathrm{obj}} ))$. In Sec. \ref{sec:reward_abl}, we also ablate against stability rewards from prior work~\cite{zhang2024catch}, simpler distance-based rewards, and contact-based rewards. However, these have lower performance across our diverse evaluation objects, pictured in Fig. \ref{fig:simobjs}.

We supplement our grasping reward with object-hand distance and object height task rewards to aid exploration. We additionally use common penalty rewards for humanoid tasks to regularize robot behavior (self-collisions; joint velocity, acceleration, torque; joint position, velocity, and torque limits; default joint position displacement; etc)~\cite{ben2024homie, omnih2o}. Equations for reward terms are available in Appendix \ref{appendix:rewards}. 

\subsection{Curricula}
\label{sec:curricula}

\begin{figure}[t]
    \centering
    \includegraphics[width=\linewidth]{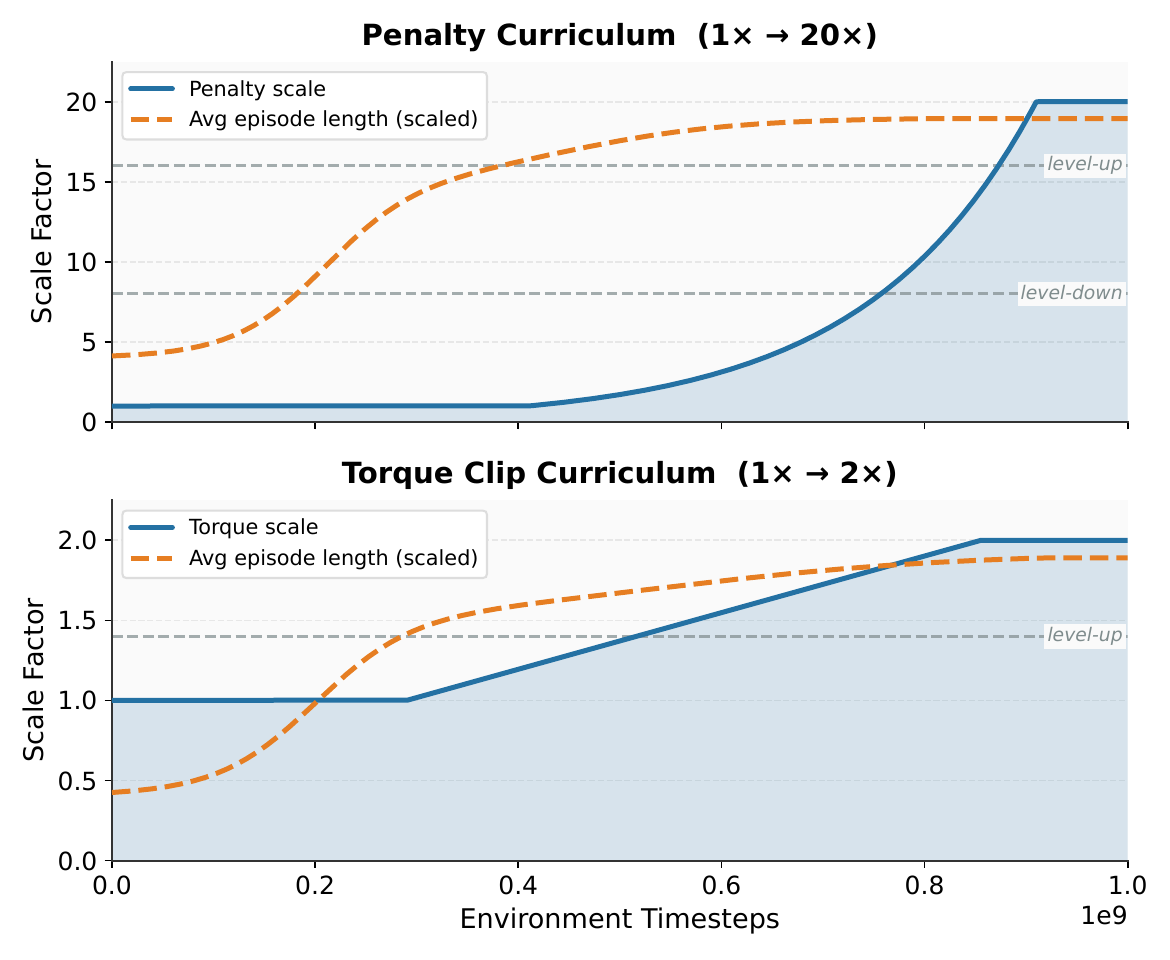}
    \vspace{-0.2in}
    \caption{\textbf{Curricula in RL training.} Penalty and reward curricula progression during RL training.}
    \label{fig:curricula}
\end{figure}

In order to improve safety and performance, we introduce three curricula used during RL policy training: (1) toss target curriculum (overhand only), (2) torque clip curriculum (underhand and overhand), and (3) penalty curriculum (underhand and overhand). The progression of the torque clip and penalty curricula during training is depicted in Fig. \ref{fig:curricula}.

\textbf{Toss target curriculum.} Learning overhand catching via online exploration is particularly difficult. To address this, we implement a toss target curriculum for overhand catches: initially, objects are thrown exactly \SI{0.1}{m} above and \SI{0.1}{m} behind the hand initial position, encouraging upright hand orientation when catching. As training progresses, the toss target range is iteratively expanded to fill the complete range of feasible overhand catches. 

\textbf{Torque clip curriculum.} With the low torque limits of the Unitree G1 arm (as low as \SI[inter-unit-product=\ensuremath{{\cdot}}]{5}{\newton\meter}), naive torque clipping yields policies that output torques exceeding hardware limits. To address this, we propose a torque clipping curriculum: initially, torques are clipped at nominal torque limits to enable effective exploration. As training progresses, the torque clip limits increase to $2\times$ nominal torque limits. Torques beyond nominal limits remain penalized in the reward, allowing the policy to explore effectively near actuator limits. Per Sec. \ref{sec:robust_and_gen} and \ref{sec:eval_tcc}, the torque clip curriculum improves performance while exhibiting lower joint mechanical power over time.

\textbf{Penalty curriculum.} Following~\cite{HumanoidVerse, omnih2o}, we implement a penalty curriculum which scales arm and hand-related regularization penalties from $1\times$ to $20\times$. Combined with the torque clip curriculum, this yields progressively greater penalties for actions beyond nominal torque limits, ensuring safer behavior.

\begin{figure*}[t!]
    \centering
    \includegraphics[width=\linewidth]{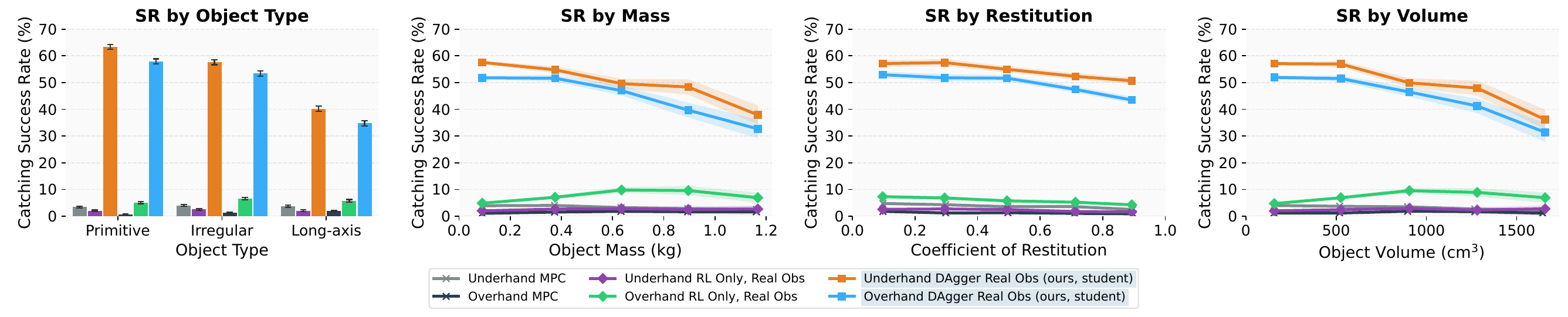}
    \vspace{-0.2in}
    \caption{\textbf{Robustness and generalization.} We test generalization to new object geometries and robustness to mass, volume, and restitution randomization. We evaluate underhand and overhand variants of the MPC, RL-only, and teacher-student distillation methods. In RL, primitive objects are seen during training, while irregular and long-axis objects are novel and unseen during training. 95\% CIs are shown via error bars and shaded regions; some CIs may be too small to see. Our method is highlighted blue in the legend. Best viewed zoomed.}
    \label{fig:type_mass_rest_vol_eval}
\end{figure*}

\section{EXPERIMENTS}
\label{sec:all_exps}

We evaluate our method in simulation (IsaacGym), sim-to-sim transfer (Mujoco), and real-world deployment. First, we demonstrate the generalization of our underhand and overhand policies across 20 objects pictured in Fig. \ref{fig:simobjs}, with variable geometry and randomized mass, elasticity, and volume. Second, we show our policies are robust to different contact dynamics, friction, and collision response via sim-to-sim in Mujoco, which enables larger-scale evaluations than is possible with real human-thrown tests. Third, we demonstrate that the torque clip curriculum yields policies which execute safer actions during catching rollouts. Fourth, we ablate the grasp reward term, comparing our per-finger distance-based grasp reward to other distance and contact-based alternatives in sim and sim-to-sim evaluations. Finally, we validate our setup by transferring zero-shot to the real world, testing underhand and overhand catching using novel objects with varying shape, deformability, and elasticity.

\subsection{Simulation and Sim-to-Sim Transfer Evaluations}
\label{sec:sim_evals}


\begin{table}[t]
\centering
\caption{\textbf{Sim and sim-to-sim evaluations.} Success rate (\%) of all policy types, 95\% CIs. Highest success rate per column is bolded; second-highest is underlined; our method is highlighted blue.}
\label{tab:robust_and_gen_sim2sim}
\resizebox{\linewidth}{!}{%
\begin{tabular}{lcccc}
\toprule
Method & \multicolumn{2}{c}{Sim SR (\%) $\uparrow$} & \multicolumn{2}{c}{Sim-to-Sim SR (\%) $\uparrow$} \\
\cmidrule(lr){2-3}\cmidrule(lr){4-5}
& Underhand & Overhand & Underhand & Overhand \\
\midrule
\multicolumn{5}{c}{\textit{Ground Truth Object Position Obs}} \\
\midrule

MPC & 3.7 $\pm$ 0.2 & 1.3 $\pm$ 0.1 & 0.1 $\pm$ 0.4 & 0.7 $\pm$ 0.6 \\

\begin{tabular}[c]{@{}l@{}}
RL G.T. Obs \\[-2pt]
w/o Torque Clip Curriculum
\end{tabular}
& \underline{75.2 $\pm$ 0.5}
& \underline{51.4 $\pm$ 0.5}
& 43.8 $\pm$ 2.7
& 8.8 $\pm$ 1.7 \\

\rowcolor{oursrow}
RL G.T. Obs (ours, teacher)
& \textbf{79.5 $\pm$ 0.4}
& \textbf{60.0 $\pm$ 0.5}
& \textbf{81.6 $\pm$ 2.2}
& \underline{22.1 $\pm$ 2.4} \\
\midrule

\multicolumn{5}{c}{\textit{Realistic Object Position Obs (Noisy, Dropped Outside Cam Frame)}} \\
\midrule

RL Only, Real Obs & 2.2 $\pm$ 0.2 & 5.8 $\pm$ 0.3 & 0.1 $\pm$ 0.4 & 0.9 $\pm$ 0.7 \\

\rowcolor{oursrow}
\begin{tabular}[c]{@{}l@{}}
DAgger Real Obs \\[-2pt]
(ours, student)
\end{tabular}
& 54.4 $\pm$ 0.5
& 49.4 $\pm$ 0.5
& \underline{56.3 $\pm$ 2.7}
& \textbf{37.0 $\pm$ 2.7} \\
\bottomrule
\end{tabular}%
}
\end{table}



As discussed in Sec \ref{sec:policy_training}, policies are trained in IsaacGym on a limited set of 7 primitive objects. Then, policies are evaluated in sim (IsaacGym) and sim-to-sim (Mujoco) on a total of 20 objects, pictured in Fig. \ref{fig:simobjs}: the training set of 7 primitive objects, a held-out set of 7 irregular objects with unseen geometries, and a held-out set of 6 long-axis objects with high angular velocity.

\subsubsection{Experimental Setup}
\label{sec:exp_setup}
We evaluate 32,768 episodes per policy in IsaacGym, and 1280 episodes per policy in Mujoco. For IsaacGym evaluations, we maintain all randomizations described in Sec. \ref{sec:policy_training}. For sim-to-sim evaluations, we do not perform additional dynamics randomizations, however we do add observation noise and camera latency. Since the Unitree API does not limit torque output, torque clipping is disabled for all evaluations. See Fig. \ref{fig:simobjs} for objects used in simulation evaluation.

\subsubsection{Performance and Generalization}
\label{sec:robust_and_gen}

To evaluate performance and generalization, we evaluate five policy types:
\begin{enumerate}
    \item\label{item:mpc} \textbf{MPC, G.T. Obs}: MPC~\cite{mpc} baselines for underhand and overhand catching.
    \item\label{item:rlonlyrealobs} \textbf{RL Only, Real Obs}: RL policy trained directly on noisy/dropped object positions. As described in Sec. \ref{sec:curricula}, the torque clip and penalty reward curricula are enabled for underhand and overhand catching, and the toss target curriculum is enabled for overhand catching.
    \item\label{item:rlgtobsnotcc} \textbf{RL, G.T. Obs w/o Torque Clip Curriculum}: RL policy trained on ground-truth (G.T.) object positions, with the torque clip curriculum disabled. The remaining curricula are used as in \rlonlyrealobs.
    \item\label{item:rlgtobsteacher} \textbf{RL, G.T. Obs (ours, teacher)}: RL policy trained on ground-truth (G.T.) object positions. All curricula are enabled as in \rlonlyrealobs. This policy is used as the teacher for the sim-to-real deployed student policy.
    \item\label{item:daggerrealobsstudent} \textbf{DAgger, Real Obs (ours, student)}: DAgger student policy distilled from teacher policy \rlgtobsteacher. During student training, all curricula and torque clipping are disabled. This is the policy used for sim-to-real deployment.
\end{enumerate}

\textbf{Does our method show strong generalization to unseen objects with variable physical properties?} In Fig. \ref{fig:type_mass_rest_vol_eval}, we evaluate the distilled student policy trained under our method on objects of variable geometry, mass, restitution, and volume. Here, the G.T. teacher policy is trained with our grasp reward, and with all curricula enabled. The torque-clip curriculum and grasp reward formulation are ablated further in Sec. \ref{sec:eval_tcc} and Sec. \ref{sec:reward_abl} respectively.

Baselines \mpc (MPC) and \rlonlyrealobs (RL only, real obs) are ineffective, with near-zero success rates, significantly lower than our final distilled policy. Moreover, while our policies are only trained on primitive objects, they generalize well to irregular object types with a performance degradation of approximately $5\%$. However, performance degrades by approximately $20\%$ on long-axis objects, which have high angular velocity not captured by our state parameterization.

Additionally, our method exhibits minimal degradation across restitution coefficients, indicating the learned grasps and movements are robust to highly elastic objects. Moreover, while performance degrades by approximately $20\%$ for very large mass and volume, performance only degrades by approximately $5\%$ for masses up to \SI{0.6}{kg} and volume up to \SI{1000}{cm^3}. Overall, our method shows strong generalization to unseen irregular objects with variable physical properties.

\textbf{Is our method robust to novel robot and contact dynamics?} In Table \ref{tab:robust_and_gen_sim2sim}, we evaluate all training methods in sim (IsaacGym) and sim-to-sim (Mujoco) to evaluate robustness to novel contact dynamics, friction, and collision response at a scale larger than is possible with real human-thrown tests. Here, baselines \mpc and \rlonlyrealobs again perform poorly in both sim and sim-to-sim transfer.

While baseline \rlgtobsnotcc (RL G.T. obs, w/o torque clip curriculum) achieves similar performance to \rlgtobsteacher (our method's RL teacher policy) in sim evaluations, its performance drops significantly in sim-to-sim evaluations. This indicates the torque clip curriculum is important for learning actions robust to novel dynamics.

Meanwhile, both our teacher policy \rlgtobsteacher and our distilled student policy \daggerrealobsstudent{} outperform all other baselines in sim-to-sim transfer, demonstrating our method's robustness to novel contact and robot dynamics.

\begin{figure}[t]
    \centering
    \includegraphics[width=\linewidth]{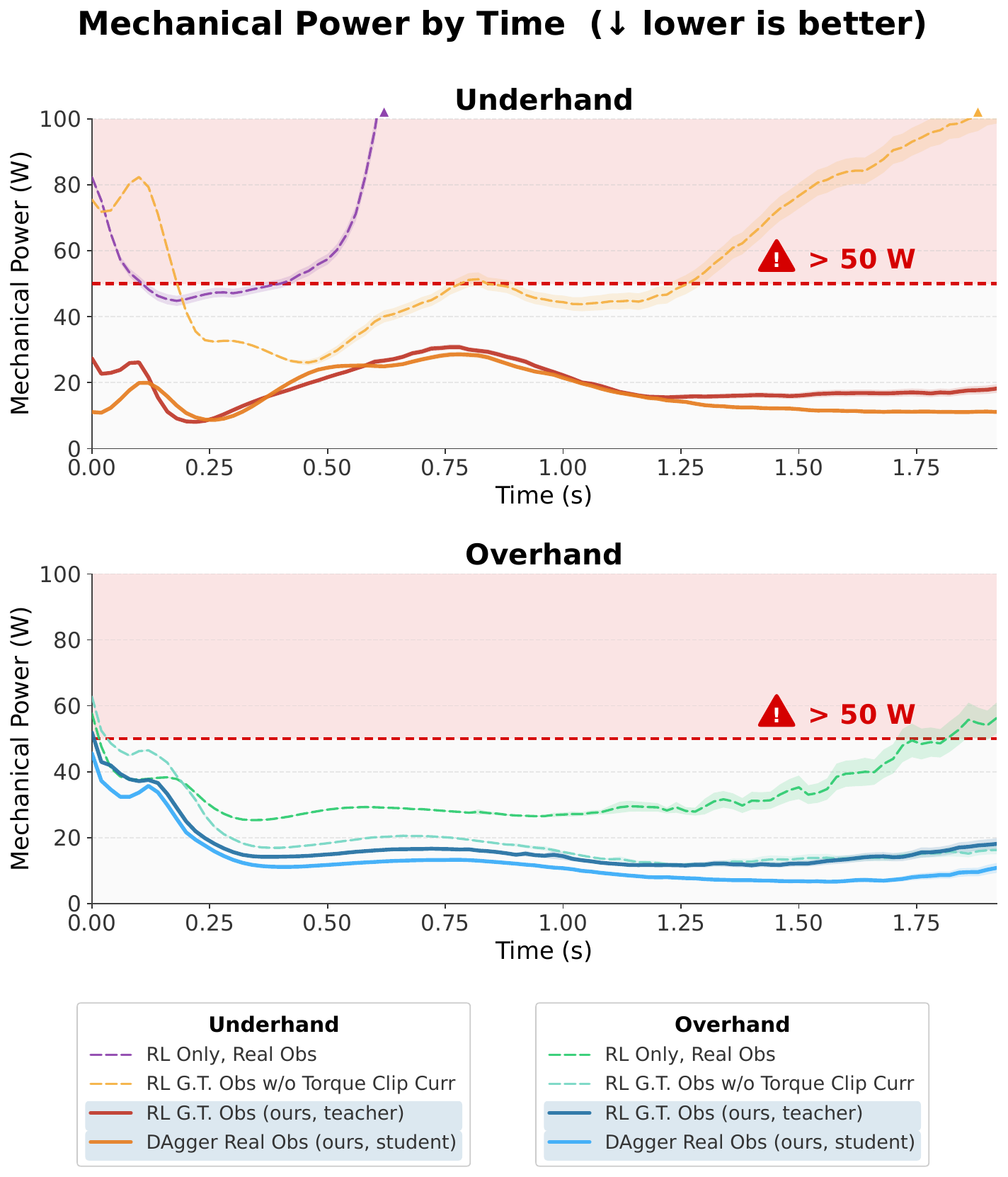}
    \vspace{-0.2in}
    \caption{\textbf{Joint mechanical power.} We plot mechanical power summed across joints, averaged over 32,768 episodes per policy. 95\% CIs are shown via shaded regions; some CIs may be too small to see. Our method is highlighted blue in the legend. Best viewed zoomed.}
    \label{fig:mechanical_power}
\end{figure}

\subsubsection{Safe Action Execution via Torque Clip Curriculum}
\label{sec:eval_tcc}

\textbf{Does our proposed torque clip curriculum yield safer action execution?} As discussed in Sec. \ref{sec:robust_and_gen}, the torque clip curriculum is important for robustness to novel dynamics in sim-to-sim transfer. Here, we additionally investigate the torque clip curriculum's impact on safe action execution. In Fig. \ref{fig:mechanical_power} we plot joint mechanical power summed across joints per timestep for baselines \rlonlyrealobs-\daggerrealobsstudent{} in both underhand and overhand catching.

Although baseline \rlonlyrealobs is trained with the torque clip curriculum, it exhibits the highest joint mechanical power, likely due to its poor overall performance as discussed in Sec. \ref{sec:robust_and_gen}. Meanwhile, although baseline \rlgtobsnotcc achieves similar success rate as our teacher policy \rlgtobsteacher in sim evaluation, it exhibits significantly higher joint mechanical power in underhand catching, and slightly higher in overhand catching. Finally, our teacher policy \rlgtobsteacher trained with the torque clip curriculum exhibits lower joint mechanical power than the baselines, as does the distilled student policy \daggerrealobsstudent.

Policies trained with the torque clip curriculum (or distilled thereof) demonstrate robustness to novel dynamics and exhibit lower joint mechanical power, demonstrating that our proposed curriculum enables safe, high-frequency action execution during catching.

\begin{table}[t]
\centering
\caption{\textbf{Grasp reward ablation.} Sim and sim-to-sim success rate (\%), 95\% CIs. Highest success rate per column is bolded; second-highest is underlined; our method is highlighted blue.}
\label{tab:reward_ablation_sim2sim}
\resizebox{\linewidth}{!}{%
\begin{tabular}{lcccc}
\toprule
Reward & \multicolumn{2}{c}{Sim SR (\%) $\uparrow$} & \multicolumn{2}{c}{Sim-to-Sim SR (\%) $\uparrow$} \\
\cmidrule(lr){2-3}\cmidrule(lr){4-5}
& Underhand & Overhand & Underhand & Overhand \\
\midrule
Object-Palm Distance Only & \underline{73.8 $\pm$ 0.5} & \underline{58.5 $\pm$ 0.5} & 16.2 $\pm$ 2.1 & 8.2 $\pm$ 1.6 \\
Object-Palm Stability~\cite{zhang2024catch} & 17.1 $\pm$ 0.4 & 31.2 $\pm$ 0.5 & \underline{18.0 $\pm$ 2.2} & \underline{10.2 $\pm$ 1.8} \\
Fingertip/Palm-Object Contact & 9.0 $\pm$ 0.3 & 22.9 $\pm$ 0.5 & 5.8 $\pm$ 1.4 & 3.4 $\pm$ 1.1 \\

\rowcolor{oursrow}
\begin{tabular}[c]{@{}l@{}}
Fingertip-Object \\[-2pt]
Surface Distance (ours)
\end{tabular}
& \textbf{79.5 $\pm$ 0.4}
& \textbf{60.0 $\pm$ 0.5}
& \textbf{81.6 $\pm$ 2.2}
& \textbf{22.1 $\pm$ 2.4} \\

\begin{tabular}[c]{@{}l@{}}
Fingertip Distance +\\[-2pt]
Fingertip/Palm-Object Contact
\end{tabular}
& 33.1 $\pm$ 0.5
& 31.1 $\pm$ 0.5
& 15.3 $\pm$ 2.1
& 1.1 $\pm$ 0.7 \\
\bottomrule
\end{tabular}%
}

\end{table}

\subsubsection{Reward Ablation}
\label{sec:reward_abl}

\textbf{Does our proposed fingertip-object surface reward improve catching performance?} We compare five reward terms for learning effective capture during catching, including one proposed by~\cite{zhang2024catch}. Each policy utilizes all curricula as described in Sec. \ref{sec:curricula}. Object-palm distance, object height, and penalty terms described in Sec. \ref{sec:rewards} are kept the same. The default reward scale for the following terms is $4.0$:
\begin{enumerate}
    \item \textbf{Object-Palm Distance Only}: Reward increases as distance between the object and palm decreases. Since this reward term is included by default with a reward scale of 4.0, we simply increase the reward scale to 8.0 (see Appendix \ref{appendix:rewards} for details on default rewards): $\exp(-2\|p_{\mathrm{obj}}-p_{\mathrm{palm}}\|)$
    \item \textbf{Object-Palm Stability}: We incorporate the object stability reward from~\cite{zhang2024catch}, which increases with the cumulative time the object is in the hand: $\Delta t/50.0$.
    \item \textbf{Fingertip/Palm-Object Contact}: As discussed in Sec. \ref{sec:policy_training}, our distance-based reward is designed to support grasps where all fingers may not fully reach the target object (e.g. irregular geometry, varying size, etc). To ensure a fair comparison, we design our contact-based rewards similarly: first, reward increases for each finger in contact with the object up to 3 fingers. Second, if at least two fingers contact the object, a bonus is given for palm contact as well: $0.1 \cdot \mathbb{I}[n_{\mathrm{finger}} \ge 1] + 0.2 \cdot \mathbb{I}[n_{\mathrm{finger}} \ge 2] + 0.4 \cdot \mathbb{I}[n_{\mathrm{finger}} \ge 3] + 0.3 \cdot \mathbb{I}[n_{\mathrm{finger}} \ge 2] \cdot \mathbb{I}[\text{palm contact}]$
    \item \textbf{Fingertip-Object Surface Distance (ours)}: As described in Sec. \ref{sec:rewards}, this reward increases as the maximum distance from any finger to the object surface decreases: $\exp(-10 ( \max\limits_{\mathrm{ftip}} \|p_{*}-p_{\mathrm{obj}}\|-r_{\mathrm{obj}} ))$
    \item \textbf{Fingertip Distance + Fingertip/Palm-Object Contact}: Reward terms \textbf{(3)} and \textbf{(4)} are combined, each given an equal weight of 2.0.
\end{enumerate}

Per Table \ref{tab:reward_ablation_sim2sim}, naive object-palm distance demonstrates strong performance in sim, but poor transfer sim-to-sim. Contact-based rewards demonstrate poor performance in sim and sim-to-sim evaluation, even combined with our proposed fingertip-object surface distance reward. The object-palm stability term from~\cite{zhang2024catch} demonstrates the second-strongest sim-to-sim transfer. However, our fingertip-object surface distance reward outperforms all baselines in both sim and sim-to-sim evaluations. This demonstrates the effectiveness and necessity for our proposed grasp reward for improving catching performance.

\subsection{Real-World Experiments}
\label{sec:real_world}

\begin{table}[t]
\centering
\caption{\textbf{Real-world evaluations.} Underhand and overhand catch success rate (\%).}
\label{tab:object_results}
\small
\setlength{\tabcolsep}{5pt}
\renewcommand{\arraystretch}{1.1}
\begin{tabular}{lcc}
\toprule
Object & Underhand SR (\%) $\uparrow$ & Overhand SR (\%) $\uparrow$ \\
\midrule
Plush           & 50.0  & 33.3 \\
Deformable      & 25.0  & 41.7 \\
Rigid-Elastic    & 41.7  & 33.3 \\

\midrule
\textbf{All Objects} & \textbf{38.9} & \textbf{36.1} \\
\bottomrule
\end{tabular}
\end{table}

\begin{figure}[t]
    \centering
    \includegraphics[width=\linewidth]{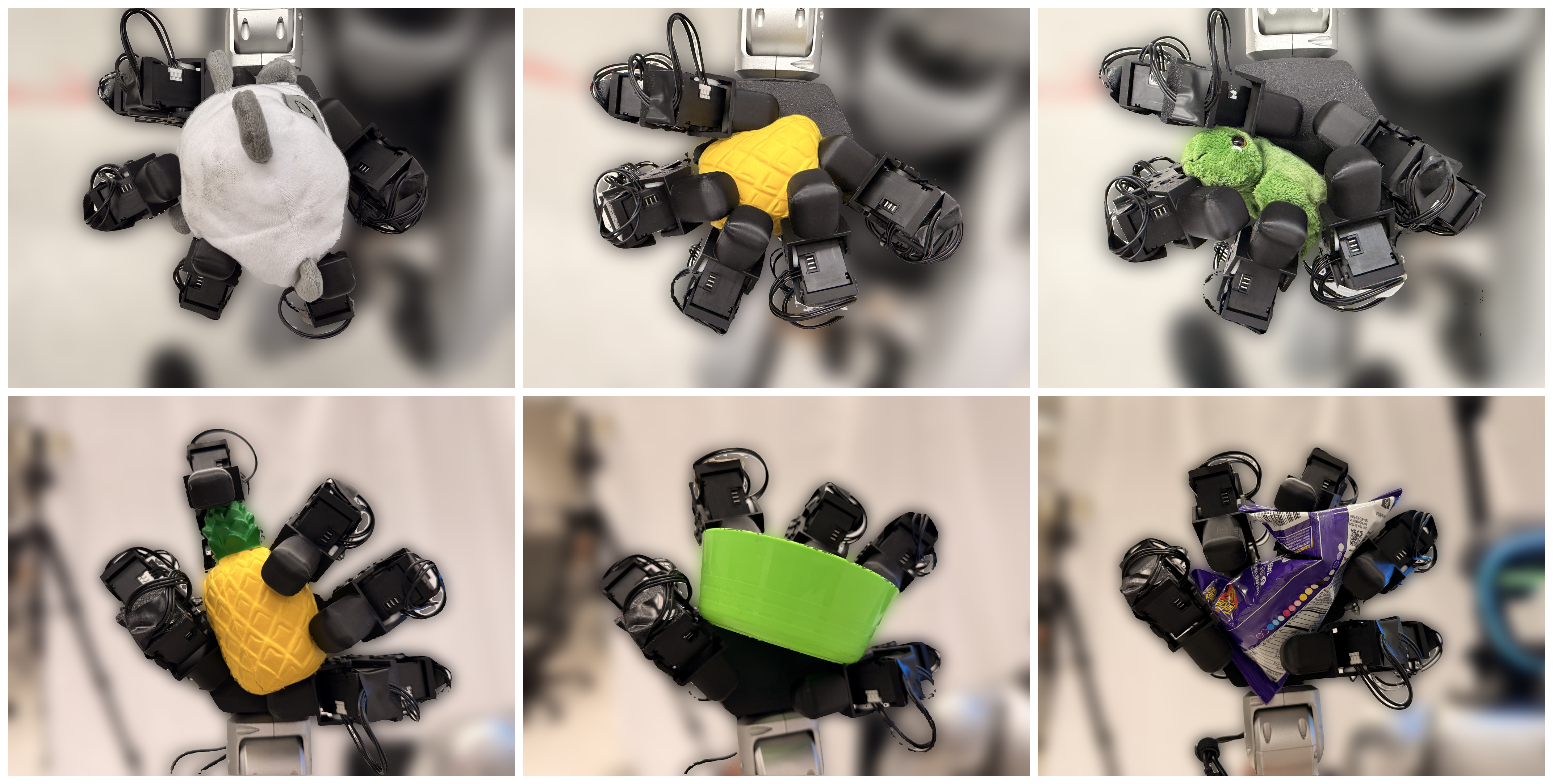}
    \vspace{-0.2in}
    \caption{\textbf{Qualitative Grasp Examples.} Examples of successful underhand and overhand grasps across different objects from real-world catching trials. Grasps adjust to object geometry and pose in-hand, yielding secure grasps.}
    \label{fig:strong_grasps}
\end{figure}

We validate the generalization capability of our method through real-world underhand and overhand catches using objects unseen during training, pictured in Fig. \ref{fig:coverish}. To ensure variety, we choose objects across three major categories: 

\begin{enumerate}
    \item \textbf{Plush (panda, turtle, sandbag)}: low rebound on contact, like sandbags used in prior work~\cite{huang2023dynamic, zhang2024catch}.
    \item \textbf{Deformable (Takis, strawberry, pineapple)}: deformables are unsupported by rigid-body simulators, and are thus novel objects for our policies.
    \item \textbf{Rigid-elastic (green bowl, ellipsoid, bear)}: these objects often rebound off rigid surfaces in unpredictable ways; we include these to test grasp quality and object capture ability.
\end{enumerate}

For each real object in Fig. \ref{fig:coverish}, we perform 1 long (\SI{>1.8}{m}), 2 medium (\SIrange{1.4}{1.8}{m}), and 1 short (\SIrange{1.0}{1.4}{m}) toss. Shorter tosses require fast reaction time, while longer tosses require effective tracking. As seen in Table \ref{tab:object_results}, we achieve a catching success rate of $38.9\%$ underhand and $36.1\%$ overhand across a diverse set of objects, demonstrating our method's strong sim-to-real transfer capability and generalizability. See Fig. \ref{fig:strong_grasps} for qualitative examples of real-world underhand and overhand post-catch grasps in the real world across objects.

\textbf{Infeasibility of teleoperated data collection in real world.} We provide a teleoperated catching baseline using a bi-manual setup~\cite{jiang2026cross} based on Homie~\cite{ben2024homie} which runs at 15 Hz. Even with an expert teleoperator, only 3/44 catches were successful, a 6.8\% success rate. The high latency and low control frequency of teleoperation make real data collection infeasible for such dynamic tasks, motivating our choice of online RL with high-throughput simulation.

\section{CONCLUSION}

We present \textit{Catch It All!}, a framework for dynamic underhand and overhand catching with a dexterous hand, using one egocentric camera. Through careful reward and curriculum design, we learn dexterous catching policies which generalize to unseen objects with diverse physical properties while operating safely near hardware limits. However, we do not apply our framework to whole-body control. We plan to extend our framework for dynamic dexterous manipulation near hardware limits to whole-body humanoid control, increasing the workspace and potentially improving system performance as a result.




\section*{APPENDIX}

\begin{table}[h!]
\centering
\caption{\textbf{Reward equations.} RL teacher catching policy.}
\label{tab:reward_table}
\setlength{\tabcolsep}{4pt}
\renewcommand{\arraystretch}{1.05}
\small
\setlength{\tabcolsep}{1pt}
\begin{tabular}{@{}lcc@{}}
\toprule
Reward Term & \textbf{Equation} & \textbf{Scale} \\
\midrule
\multicolumn{3}{c}{\textit{Task}} \\
\midrule
\multirow{3}{*}{\shortstack[l]{Approx\\Fingertip-Obj\\Surface Dist}} & \multirow{3}{*}{$\exp(-10 ( \max\limits_{\mathrm{ftip}} \|p_{*}-p_{\mathrm{obj}}\|-r_{\mathrm{obj}} ))$} & \multirow{3}{*}{$4.0$} \\
& & \\
& & \\[4.0pt]
Obj-Palm Dist & $\exp(-2\|p_{\mathrm{obj}}-p_{\mathrm{palm}}\|)$ & $4.0$ \\[4.0pt]
Obj Height & $\mathrm{clip}(p^z_{\mathrm{obj}}-0.2,0.0,1.3) / 1.3$ & $1.0$ \\
\midrule
\multicolumn{3}{c}{\textit{Posture / Contact}} \\
\midrule
DoF Default Pos & $\|q-q_{\mathrm{default}}\|_1$ & $-1.0$ \\
Body Collision & $\|F_{\mathrm{body}}\|_1$ & $-2.0$ \\
Arm Collision & $\|F_{\mathrm{arm}}\|_1$  & $-1.0$ \\
\midrule
\multicolumn{3}{c}{\textit{Joint Regularization}} \\
\midrule
DoF Track Error & $\|q-q_{\mathrm{target}}\|^2$ & $-0.1$ \\
Action Rate & $\|a_t-a_{t-1}\|^2$ & $-0.01$ \\
DoF Smoothness  & $\|a_t-2a_{t-1}+a_{t-2}\|^2$ & $-0.05$ \\
DoF Power & $\||\dot q|\odot|\tau|\|_1$ & $-2e{-}5$ \\
DoF Vel & $\|\dot q\|^2$ & $-1e{-}4$ \\
DoF Acc & $\|\ddot q\|^2$ & $-2.5e{-}7$ \\
Torques & $\|\tau\oslash K_p\|^2$ & $-2.5e{-}6$ \\
DoF Pos Lims & $\|(q_{\mathrm{min}}\!-\!q)_+ + (q\!-\!q_{\mathrm{max}})_+\|_1$ & $-2.0$ \\
DoF Vel Lims & $\|[|\dot q|- 0.8\dot q_{\mathrm{max}}]_+\|_1$ & $-2e{-}3$ \\
Torque Lims & $\|[|\tau|-0.95\tau_{\mathrm{max}}]_+\|_1$ & $-0.1$ \\
Action Vanish & $\|[a-a^{\max}]_+ + [a^{\min}-a]_+\|_1$ & $-1.0$ \\
\midrule
\multicolumn{3}{c}{\textit{Hand Regularization}} \\
\midrule
Hand Pos Rate  & $\|p^\mathrm{hand}_{t}-p^\mathrm{hand}_{t-1}\|$ & $-0.05$ \\
Hand Smoothness & $\|p^\mathrm{hand}_{t}-2p^\mathrm{hand}_{t-1}+p^\mathrm{hand}_{t-2}\|$ & $-0.05$ \\
Hand Orientation & $\|g^{xy}_{\mathrm{hand}}\|^2$ & $-0.025$ \\
\bottomrule
\end{tabular}
\end{table}

\subsection{Reward Equations}
\label{appendix:rewards}

Table \ref{tab:reward_table} lists all rewards, using the following variables:

\noindent
\begin{tabular}{@{}r@{\hspace{6pt}}l@{}}
$p_{\mathrm{item}}$ & Cartesian position of $\mathrm{item}$ (object, link, etc.) \\
$r_{\mathrm{item}}$ & (Approximate) radius of $\mathrm{item}$ \\
$F_{\mathrm{item}}$ & Force applied to $\mathrm{item}$ \\
$q, \dot{q}, \ddot{q}$ & Joint position, velocity, acceleration \\
$q_{\mathrm{target}}$ & Controller joint target \\
$q_{\mathrm{default}}$ & Default joint positions \\
$a_t$ & Action at timestep $t$ \\
$\tau$ & Torques \\
$x_{\mathrm{min}}, x_{\mathrm{max}}$ & Min/max values for quantity $x$ \\
$g_{\mathrm{item}}$ & Projected gravity vector of $\mathrm{item}$ \\
$K_p, K_d$ & PD Gains
\end{tabular}

\bibliographystyle{IEEEtran}
\bibliography{references}

\end{document}